# Adaptive Degradation Process with Deep Learning-Driven Trajectory

Li Yang, Beihang University, yanglirass@buaa.edu.cn

*Abstract*— Remaining useful life (RUL) estimation is a crucial component in the implementation of intelligent predictive maintenance and health management. Deep neural network (DNN) approaches have been proven effective in RUL estimation due to their capacity in handling high-dimensional non-linear degradation features. However, the applications of DNN in practice face two challenges: (a) online update of lifetime information is often unavailable, and (b) uncertainties in predicted values may not be analytically quantified. This paper addresses these issues by developing a hybrid DNN-based prognostic approach, where a Wiener-based-degradation model is enhanced with adaptive drift to characterize the system degradation. An LSTM-CNN encoder-decoder is developed to predict future degradation trajectories by jointly learning noise coefficients as well as drift coefficients, and adaptive drift is updated via Bayesian inference. A computationally efficient algorithm is proposed for the calculation of RUL distributions. Numerical experiments are presented using turbofan engines degradation data to demonstrate the superior accuracy of RUL prediction of our proposed approach.

## I. Introduction

Recent years have witnessed the prominent advancements of machine learning technologies, particularly deep neural network (DNN) approaches in the arsenal of prognostics and health management (PHM) [1]. In contrast to the other data-driven prognostic approaches, DNN has the advantage of enabling automatically high-level feature extractions [2]. Common DNN structures include Restricted Boltzmann Machine (RBM) [3], Recurrent Neural Network (RNN) [4, 5], Convolutional Neural Networks (CNN) [6, 7], which have been successfully employed for predictions of remaining useful lifetime (RUL) with good accuracy performance.

Despite the prominent success of DNN, its application in health prognostics faces two challenges. The first is the generalization problem, as the performance of DNN could be poor when processing previously unseen sensor data [8]. This is often the case in actual asset health monitoring, where monitoring data are collected from a limited number of machines. Consequently, prognostic capacity is limited, posing a challenge to DNN-based prognostics when new mechanisms emerge. Second, prediction uncertainty, a common problem in RUL estimation, is not well addressed by most DNN approaches, which may mislead maintenance decision-makings and result in production losses [9]. Common uncertainty sources include: (a) unavoidable random factors, such as measurement error and interference noise (aleatoric uncertainty), and (b) ignorance of prognostic model (epistemic uncertainty), which is associated with data size [10].

A potential solution to the above problems is to characterize health conditions via degradation models [11]. Stochastic processes such as Wiener process are commonly used in degradation models for RUL prediction [12-14]. On one hand, prognostic results under such model-driven frameworks are reflected by density functions, which capture aleatoric uncertainties. On the other hand, recent advancements of Wiener-based RUL allow for adaptive drift rates so that unit-to-unit variability can be addressed and online update can be implemented [15-18]. This is particularly useful in alleviating generalization problem and quantifying epistemic uncertainties. However, compared with data-driven approaches, such model-driven methods are relatively incapable of learning multiple failure patterns and extracting hidden patterns from historical data. Moreover, the assumption that degradation follows some pre-set trajectories limits their actual application since: (a) pre-set trajectory is based on the statistical averaging idea, which is not suitable to learning comprehensive degradation patterns from historical data, and (b) pre-set trajectory may mislead prognostic results under complex environmental stresses and varying operation conditions.

In this paper, we propose a novel hybrid prognostic framework in which deep learning-based trajectory prediction enables the degradation model to overcome the limitations of pre-set trajectories of the traditional degradation models. In the proposed framework, the non-linear Wiener process with adaptive drift is chosen as the degradation model, while the future degradation trajectory (shape) is trained via an encoder-decoder [19] as a combination of Long Short-Term Memory (LSTM) and Convolutional Neural Network (CNN) to predict multi-step trajectory. As such, our Wiener-based degradation model is capable of extracting hidden degradation patterns from historical data. At the same time, the degradation rate can be dynamically updated as new sensor data becomes available, which further enhances the generalization of DNN model. To the best of our knowledge, this is the first attempt to integrating adaptive learning abilities into stochastic degradation models, significantly enhancing their compatibility and flexibility to sensor data.

In addition to the unique combination of degradation model and encoder-decoder neural network, the framework involves computationally efficient algorithms to quantify the uncertainties in RUL prediction. At the model training stage, evolution patterns and noise degrees are jointed learned via a single DNN structure, which formulates variance terms as a new network branch. At the RUL density calculation stage, an interpolation approximation algorithm is developed for the generation of virtual degradation paths. Unlike Particle Filter

(PF) [20], the proposed algorithm can avoid computationally expensive iterative simulation, and allows the generation of the entire virtual paths in one step employing prior knowledge of Gaussian distribution.

In summary, this paper makes the following contributions to the existing prognostic literature:
1) A hybrid prognostic framework integrating Wiener-based degradation model and deep neural network (DNN) is formulated, which captures prediction uncertainties with real-time updates;
2) The future trajectory of the Wiener model is learned via an LSTM-CNN encoder-decoder, which better reflects the variations of operation/health conditions in a more sensitive and flexible way;
3) An extended DNN structure is proposed to simultaneously learn the degradation trajectory and the variances of the Wiener processes in a single neural network;
4) An interpolation approximation algorithm is developed for uncertainty quantification of RUL, which is more concise and computation-effective than PF.

The rest of the paper is organized as follows: Section II introduces the prognostic methodology. Section III focuses on numerical studies. Section IV discusses the effectiveness of the proposed framework. Section V concludes the paper.

## II. METHODOLOGY

Consider a continuous-time degradation process $\{Z(t), t > t_0\}$, where $Z(t)$ represents the degradation level at time $t$. The device is deemed failed when $Z(t)$ exceeds a pre-set threshold $D$. The evolution of the degradation process is controlled by a pattern term $Q(t;\Theta)$, where $\Theta$ are unknown parameters. To be specific, we can formulate $Q(t;\Theta)$ as a time-series as follows:

$$Q(t;\Theta) = f\left(Z_{t_i}, Z_{t_{i-1}}, Z_{t_{i-2}}, ..., Z_{t_0}, t; \Theta\right), t > t_i, \quad (1)$$

where $t_i$ is the measurement point prior to $t$, $Z_{t_i}$ is the observation at $t_i$, and $f$ is a nonlinear function parametrized by unknown parameters $\Theta$. Eq. (1) states that the future trajectory is dependent on both the operation time and previous degradation path. In other words, the pattern term varies over time and can be learned by analyzing sensor data. In this paper, a Wiener process with adaptive drift is used to characterize the degradation process. Accordingly, the observation function at $t$ is written as

$$Z(t) = Z_{t_i} + \int_{t_i}^{t} \varphi(\tau) dQ(\tau;\Theta) + \int_{t_i}^{t} \eta(\tau) dB(\tau)$$
$$= Z_{t_i} + \int_{t_i}^{t} \varphi(\tau) df\left(Z_{t_i}, Z_{t_{i-1}}, Z_{t_{i-2}}, ..., Z_{t_0}, t; \Theta\right) + \int_{t_i}^{t} \eta(\tau) dB(\tau), \quad t > t_i, \quad (2)$$

where $\varphi(t)$ is the adaptive drift, $\eta(t)$ is the diffusion parameter, and $B(.)$ represents a standard Brownian motion (BM). Note that $\eta(t)$ reflects the noise, which can usually be reduced to $\eta(t) = \eta_B$ due to its stability. The drift rate $\varphi(t)$ is a continuous and time-varying stochastic process, which is characterized by another Wiener process, i.e.,

$$\varphi(t) = \varphi_{t_i} + \gamma \Lambda(t - t_i), \quad (3)$$

where $\gamma$ is the diffusion parameter, and $\Lambda(.)$ represents the standard BM. Eqs. (2)-(3) summarize three key influencing factors of the degradation trajectory: (a) data-driven evolution pattern $Q(t;\Theta)$, (b) instantaneous degradation rate $\varphi(t)$ affected by random environmental disturbance, and (c) unavoidable noise degree $\eta_B$.

Many of recent advancements of adaptive BMs can be viewed as special cases of our model, with a preset and constant evolution pattern. Common forms include power law [15], exponential [16] and linear [15, 17]. However, their robustness under varying or complex operational conditions may be limited, and historical measurement data are not sufficiently used for trajectory determination in these models. In contrast, our model can extract the evolution pattern from historical data while maintaining the self-updating ability. The pattern term $Q(t;\Theta)$ can be characterized by arbitrary parametric functions $f$, such as linear regression, support vector machine and neural network etc. In this paper, **Deep Neural Network (DNN)** is chosen for its effectiveness in processing and learning knowledge from historical data. DNN can also be easily configured into multi-step trajectory prediction, which further enhances the robustness of the model.

With such deep-learning driven evolution patterns, the proposed degradation model gains both the self-updating and learning abilities. The prognostic process based on this adaptive model can be decomposed into the following three steps.
(1) Training of the degradation model via DNN;
(2) Calculation of RUL density function;
(3) Real-time update of adaptive drift.

### A. Model Training

Our model has a number of parameters to learn and uses Bayesian network as illustrated in Fig. 1, where the dark solid lines represent the probabilistic relationship and the red dashed lines are deterministic connections. Unknown degradation model parameters can be classified into two categories: (a) root nodes of the Bayesian network, including the diffusion parameter $\gamma$, the noise degree $\eta_B$, and initial drift $\varphi_{t_0}$, and (b) weights of the DNN model $\Theta$, which can be regarded as specific internal nodes. The root nodes of a Bayesian network are usually hyper-parameters and determined by experience. In contrast, our framework allows for analytical forms of the likelihood and root node values via Maximum Likelihood Estimation (MLE), thanks to nice mathematical properties of BM. The estimation steps are described below.

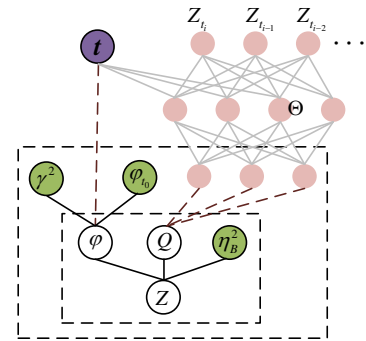

Fig. 1. Graph representation of the Bayesian network.

*1) Setting of $\varphi_{t_0}$*

According to the statistical properties of BM, the degradation observation at any measurement point $t$ follows a Gaussian distribution. Its expectation is calculated as

$$\begin{aligned}E(Z(t))&=E\left(Z_{t_i}+\int_{t_i}^{t}\varphi(\tau)dQ(\tau;\Theta)+\eta_B B(t-t_i)\right)\\&=Z_{t_i}+\int_{t_i}^{t}E(\varphi(\tau))dQ(\tau;\Theta)+0\\&=Z_{t_i}+E(\varphi(t_i))(Q(t;\Theta)-Q(t_i;\Theta))\\&=E(\varphi(t_i))Q(t;\Theta)+Z_{t_i}-E(\varphi(t_i))Q(t_i;\Theta).\end{aligned} \quad (4)$$

The term $Z_{t_i}-E(\varphi(t_i))Q(t_i;\Theta)$ is constant. As $Q(t;\Theta)$ only affects the trajectory after $t_i$, we can assign $Q(t_i;\Theta)=Z_{t_i}/E(\varphi_{t_i})$ to eliminate this term. Besides, $E(\varphi(t_i))=\varphi_{t_0}$ holds in the training stage since prior knowledge is not available. Therefore, Eq. (4) can be simplified as

$$E(Z(t))=\varphi_{t_0}Q(t;\Theta). \quad (5)$$

Eq. (5) indicates that, although the pattern term $Q(t;\Theta)$ is not observable, it is proportional to the expectation of real observations. As $Q(t;\Theta)$ is a parametric function controlled by DNN weights $\Theta$, any scaling of $\varphi_{t_0}$ can be offset by adjusting $\Theta$. Therefore, we can simply set $\varphi_{t_0}=1$, such that $Q(t;\Theta)$ amounts to the expected degradation in the training stage, i.e., $E(Z(t))=Q(t;\Theta)$.

*2) Estimation of $\gamma$, $\eta_B$ and DNN weights $\Theta$*

The noise coefficients $\gamma$, $\eta_B$ and DNN weights $\Theta$ can be jointly estimated by an extended DNN structure. According to the Bayesian theory, the variance of observation $Z(t)$ can be decomposed as

$$\begin{aligned}\text{Var}(Z(t))&=\text{Var}\left(Z_{t_i}+\int_{t_i}^{t}\varphi(\tau)dQ(\tau;\Theta)+\eta_B B(t-t_i)\right)\\&=\gamma^2\int_{t_i}^{t}(Q(t_i;\Theta)-Q(\tau;\Theta))^2 d\tau+\eta_B^2(t-t_i).\end{aligned} \quad (6)$$

Eq. (6) reveals the fundamental uncertainty structure of the degradation model. The first term in the right hand side represents the variability of drift $\varphi$ (known as epistemic uncertainty), and the second term reflects the variability from random factors such as noise, measurement error (known as aleatoric uncertainty). Therefore, Eq. (6) follows the law of the total variance [21], which captures the comprehensive effects of two common prognostic uncertainties. With the expectation obtained in Eq. (5) and variance obtained in Eq. (6), the entire likelihood for degradation observations can be formulated.

In realistic condition monitoring, sensor data are collected sequentially. Sliding window technique is employed to set the input data and output parameters for model training. Fig. 2 illustrates a basic one-step prediction, where all the sensor data between $t_0$ and $t_i$ in sequence $j$ are input for prediction of the subsequent observation $z_{t_i}^j$. Accordingly, the training pair at $t_i$ is $[z_{t_{i-1}}^j, z_{t_{i-2}}^j, ..., z_{t_0}^j] \to z_{t_i}^j$. Then new training pairs can be created iteratively.

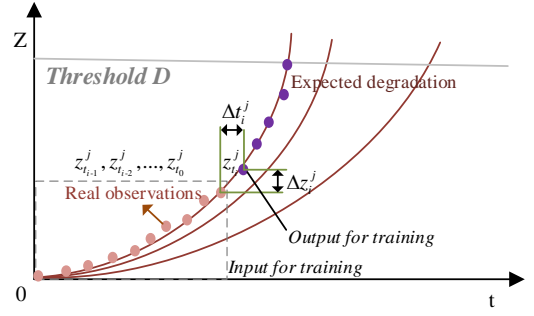

Fig. 2. Training pair illustration of one-step prediction.

According to Eq. (6), the prediction variance of $Z_{t_i}$ based on training sequence $[Z_{t_{i-1}}, Z_{t_{i-2}}, ..., Z_{t_0}]$ is

$$\text{Var}(Z_{t_i})=\gamma^2\int_{t_{i-1}}^{t_i}(Q(t_i;\Theta)-Q(\tau;\Theta))^2 d\tau+\eta_B^2(t_i-t_{i-1}). \quad (7)$$

Through one-step numerical integration [22], Eq. (7) can be further simplified as

$$\begin{aligned}\text{Var}(Z_{t_i})&\approx\gamma^2\int_{t_{i-1}}^{t_i}\left(\frac{\Delta Q_i}{\Delta t_i}\tau\right)^2 d\tau+\eta_B^2\Delta t_i\\&=\tfrac{1}{3}\gamma^2\Delta Q_i^2\Delta t_i+\eta_B^2\Delta t_i\approx\tfrac{1}{3}\gamma^2\Delta Z_i^2\Delta t_i+\eta_B^2\Delta t_i,\end{aligned} \quad (8)$$

where $\Delta Q_i = Q_{t_i}-Q_{t_{i-1}}$, $\Delta Z_i = Z_{t_i}-Z_{t_{i-1}}$, $\Delta t_i = t_i-t_{i-1}$, as illustrated in Fig. 2. This is a heteroscedasticity regression problem since the variance is not constant. Therefore, the widely accepted root-mean-square error (RMSE) cannot be used for training. Instead, Eqs. (5)-(6) allows us to formulate the whole negative log-likelihood of the Gaussian distribution (constants are ignored) as

$$L\propto\frac{1}{n}\sum_{i=1}^{n}\left(\frac{1}{2}(\text{Var}(Z_{t_i}))^{-1}(Z_{t_i}-Q(t;\Theta))^2+\frac{1}{2}\ln(\text{Var}(Z_{t_i}))\right). \quad (9)$$

Substituting Eq. (8) into Eq. (9) yields

$$L\propto\frac{1}{n}\sum_{i=1}^{n}\left(\begin{array}{l}\frac{1}{2}(\tfrac{1}{3}\gamma^2\Delta Z_i^2\Delta t_i+\eta_B^2\Delta t_i)^{-1}(Z_{t_i}-Q(t;\Theta))^2+\\ \frac{1}{2}\ln(\tfrac{1}{3}\gamma^2\Delta Z_i^2\Delta t_i+\eta_B^2\Delta t_i)\end{array}\right). \quad (10)$$

The unknown parameters of this likelihood function include diffusion coefficient $\gamma$, noise degree $\eta_B$ and weight sets of the DNN model $\Theta$. The feasible generalized least squares (FGLS) is a solution when heteroscedasticity exists in a regression problem. However, FGLS may be inconsistent and time-consuming for multi-parameter model structures [23]. Alternately, we can employ gradient descents on both noise degrees and DNN weights by extending the DNN structure, thanks to the heteroscedasticity structure addressed in Eq. (8).

The key idea is to treat the unknown variance coefficients $\gamma^2$, $\eta_B^2$ as the weights of a DNN branch, as illustrated in Fig. 3. Therefore, the resulting DNN will cover two sets of model components: (a) the evolution pattern $Q$, and (b) weights $\gamma^2$ and $\eta_B^2$. As a result, the variance $\text{Var}(Z_{t_i})$ and expectation $Q(t_i;\Theta)$ can be simultaneously learned, adopting the negative log-likelihood in Eq. (10) as the loss function. As such, the parameter estimation problem represented in Fig.1 is successfully transformed into a DNN training problem.

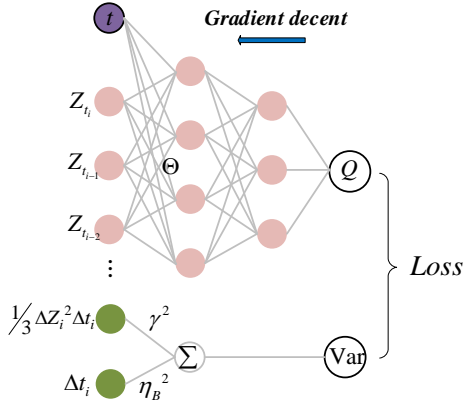

Fig.3. Structural extension of the DNN to learn the expectation and variance.

*3) Multi-step Prediction*

A challenge in time-series analysis is the instability during the long-term prediction. To alleviate this problem, it is recommended to look multiple steps ahead in a single prediction. This can be done by a DNN with multiple output nodes. Accordingly, the loss function in Eq. (9) becomes

$$L \propto \frac{1}{n}\sum_{i=1}^{n}\frac{1}{m_n}\sum_{j=0}^{m_n}\left(\begin{array}{l}\frac{1}{2}\left(\mathrm{Var}\left(Z_{t_{i+j}}\right)\right)^{-1}\left(Z_{t_{i+j}}-Q(t_{i+j};\Theta)\right)^{2}\\+\frac{1}{2}\ln\left(\mathrm{Var}\left(Z_{t_{i+j}}\right)\right)\end{array}\right), \quad (11)$$

with $n$ sequences and $m_n$ predictions for each sequence. This loss function can be seen as a mean square error weighted by $\left(\mathrm{Var}\left(Z_{t_{i+j}}\right)\right)^{-1}$. It incurs a bigger penalty for wrongly predicted points in the nearer future, as the variance is always growing. This feature is quite important when updating the posterior distribution of $v$, as we will show in Section II.C.

By Eq. (11), the multi-step prediction of $\mathrm{Var}\left(Z_{t_{i+j}}\right)$ in Eq. (7) becomes

$$\begin{aligned}\mathrm{Var}\left(Z_{t_{i+j}}\right)&=\gamma^{2}\int_{t_{i-1}}^{t_{i+j}}\left(Q(t_{i+j};\Theta)-Q(\tau;\Theta)\right)^{2}d\tau+\eta_{B}^{2}\left(t_{i+j}-t_{i-1}\right)\\&\approx\gamma^{2}\int_{t_{i-1}}^{t_{i+j}}\left(Z(t_{i+j};\Theta)-Z(\tau;\Theta)\right)^{2}d\tau+\eta_{B}^{2}\left(t_{i+j}-t_{i-1}\right),\end{aligned} \quad (12)$$

which can also be approximated via numerical integration.

*4) Structure of the DNN Model*

The proposed framework involves a LSTM-CNN encoder-decoder for multi-step prediction. The superiority of this structure comes from the following facts: (a) both inputs and outputs in multi-step predictions are multi-dimensional, to which encoder-decoder fits the best, (b) the inputs pending encoding are of variable lengths, which can be effectively handled by recurrent neural network (RNN), and (c) the prediction points in the degradation path is closely correlated to its neighboring node, to which convolutional neural network (CNN) is a more efficient structure than RNN. The encoder-decoder structure is shown in Fig.4.

As shown in Fig. 4, the running cycles are firstly normalized by the maximum cycle in the training set. Then both the historical health index (HI) and normalized running cycles are given as an input to the LSTM encoder. In this way, all the historical information is compressed into a $d$-dimensional health vector. In the decoding part, the health vector is firstly duplicated for $k$ times and then stacked with the future normalized running cycle, generating a $(d+1)\times k$ health matrix. Finally, a CNN network is used to decode the health matrix into a future degradation trajectory. Then the RUL can be determined by the first passage time to the failure threshold $D$.

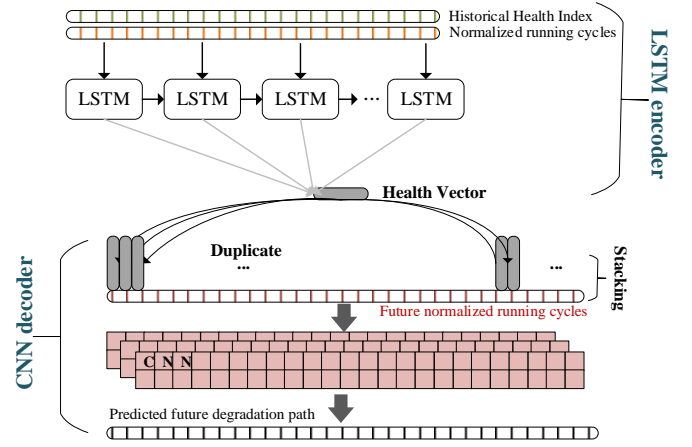

Fig. 4. Structure of the LSTM-CNN encoder-decoder.

*B. RUL Calculation*

Given the measurement data $Z_{t_0},Z_{t_1},\cdots,Z_{t_i}$, the future degradation process can be predicted by Eqs. (4) and (6). Notice that, when new sensor data becomes available at $t_i$, the expectation $\mathrm{E}(\varphi(t_i))$ in Eq. (4) is substituted by $\varphi_{t_i}$. Then the RUL calculation problem equals to the calculation of the first passage time of a stochastic process with an average trajectory $\varphi_{t_i}Q(t;\Theta)$ and time-dependent variance

$$\xi(t)=\gamma^{2}\int_{t_i}^{t}\left(Q(t_i;\Theta)-Q(\tau;\Theta)\right)^{2}d\tau+\eta_{B}^{2}\left(t-t_i\right). \quad (13)$$

Accordingly, this stochastic process can be rewritten as

$$Z(t)=\varphi_{t_i}Q(t;\Theta)+B(\xi(t)), \quad (14)$$

where $B(.)$ represents the standard Brownian motion. Due to complex non-linear evolution patterns, analytical RUL distributions under such processes are usually intractable [17]. An approximation solution for such problems via scale transformation (ST) is proposed in [14]. However, the derivatives in this approach can be numerically unstable when degradation evolutions are not smooth. Particle Filter (PF) is another possible solution [16, 18], but at the cost of expensive computation. Instead, we develop a new algorithm for RUL calculation, which can achieve satisfactory accuracy with much less computation than PF. To this end, we substitute $Z(t)$ with the following process

$$\chi(t)=\varphi_{t_i}Q(t;\Theta)+k\sqrt{\xi(t)}, \quad (15)$$

where $k \sim \mathrm{N}(0,1)$. From Eq. (15), $\chi(t)$ follows a Gaussian distribution with $\mathrm{E}(\chi(t))=\varphi_{t_i}Q(t)$ and $\mathrm{Var}(\chi(t))=\xi(t)$. Clearly, $\chi(t)$ shares the same statistical properties with $Z(t)$ and can be easily sampled. Therefore, the RUL distribution can be generated by a series of points $t_n$ where $\chi(t_n)=D$. We name the approximation algorithm as Interpolation Approximation, which is presented in Algorithm 1.

**Algorithm 1**: Interpolation Approximation of RUL

**Input**: Threshold $D$, the number of iterations $N$, degradation expectation $\varphi_{t_i}Q(t;\Theta)$, variance $\xi(t)$.

**Output**: RUL distribution $r(t)$.

**Initialization**: Empty point set $\delta$.

1: **For** $n$ in 1 to $N$, **do**
2:   Sample $k_n$ from $N(0,1)$
3:   Generate an interpolation curve
$$\chi_n(t) = \varphi_{t_i}Q(t;\Theta) + k_n\sqrt{\xi(t)},$$
4:   Calculate the cross time of $\chi_n(t)$ to $D$, i.e. $t_n = \chi_n^{-1}(D)$.
5:   Add $t_n$ into $\delta$;
6: **End for**
7: Obtain $r(t)$ from $\delta$ by kernel density estimation (KDE) with weights of $1/N$.

Fig. 5 compares the degradation trajectories generated by PF and those by the proposed interpolation algorithm. Obviously, the trajectories generated by Algorithm 1 are of higher regularity and easier to control. As an iteration-based approximation solution, PF is not only computationally expensive, but also faces the problem of degeneracy, which means all but one of the importance weights are close to zero. Sophisticated resampling techniques are required to alleviate the problem. In contrast, our proposal can avoid these difficulties since no iteration along time is needed.

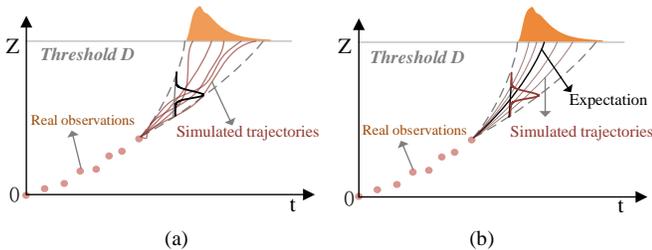

Fig. 5. Comparison between PF and the proposed algorithm: (a) PF and (b) Algorithm 1.

For validation, the performance of three algorithms (scale transformation in [15], PF and Algorithm 1) on a linear degradation process is tested. The parameters are set as $\varphi_{t_0}$ =0.01, $\gamma^2$=0.0005, $\eta_B^2$=0.0001 and $D$=0.8. Then the discrete distribution is transformed into a continuous one by kernel density estimation (KDE) with a bandwidth of 0.2. The approximation distributions with different PF particles are shown in Fig. 6. Clearly, Algorithm 1 outperforms the others by achieving a smooth distribution with only 20 interpolation lines, while PF needs 1,000 particles to gain similar results. We also find that the performance of scale transformation is limited. With 1,000 particles, the quality of distribution found by PF is close to our interpolation algorithm. The superiority of Algorithm 1 is further validated by the running time comparison in Table I. Clearly, PF requires a much longer simulation time than Algorithm 1.

In summary, our interpolation approximation algorithm has three advantages over PF. First, the entire virtual paths can be generated in one step based on prior knowledge of Gaussian distributions. As such, iteration can be avoided, which saves considerable computation. Second, less virtual paths are needed as the intersection points are more evenly distributed. Finally, discretization of degradation processes in PF is not required, so that less accuracy loss is incurred.

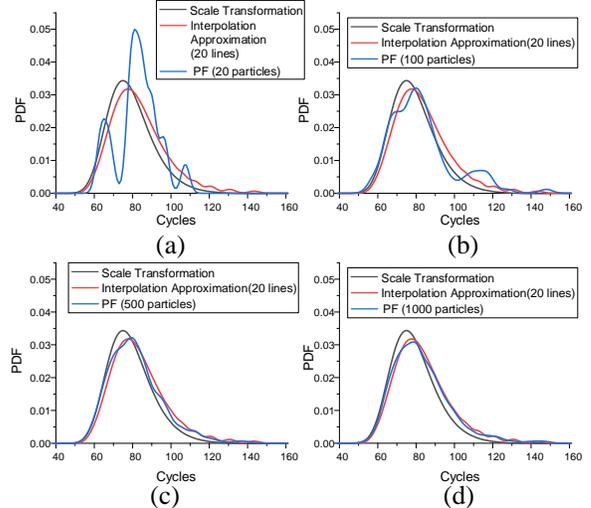

Fig. 6. The approximation RUL distribution of the scale transformation, the PF and the proposed method. a) 20 PF particles; b) 100 PF particles; c) 500 PF particles; d) 1,000 PF particles.

TABLE I
RUNNING TIME OF DIFFERENT APPROXIMATES

| Method | Algorithm 1 (20) | Scale transform | PF (20) | PF (100) | PF (500) | PF (1000) |
|---|---|---|---|---|---|---|
| Time/s | 0.011 | 0.013 | 0.030 | 0.081 | 0.325 | 0.592 |

### C. Updating of drift

The posterior inference of $v$ can be done iteratively following Bayesian update. From Eqs. (2)-(4), when the posterior distribution of $\varphi_{t_{i-1}}$ is given by $\varphi_{t_{i-1}} \sim N(\psi_{t_{i-1}}, \omega_{t_{i-1}}^2)$, the joint distribution of $\varphi_{t_i}$ and $Z_{t_i}$ follows a multi-dimensional Gaussian distribution with the following mean vector and covariance matrix

$$E\begin{pmatrix}\varphi_{t_i}\\Z_{t_i}\end{pmatrix} = \begin{pmatrix}\psi_{t_{i-1}}\\\psi_{t_{i-1}}Q(\tau;\Theta)\end{pmatrix}, \text{Cov}\begin{pmatrix}\varphi_{t_i}\\Z_{t_i}\end{pmatrix} = \begin{bmatrix}\text{Cov}_{t_i}(1,1), & \text{Cov}_{t_i}(1,2)\\\text{Cov}_{t_i}(2,1), & \text{Cov}_{t_i}(2,2)\end{bmatrix}, \quad (16)$$

where

$$\text{Cov}_{t_i}(1,1) = \gamma^2\Delta t_i + \omega_{i-1}^2,$$

$$\text{Cov}_{t_i}(1,2) = \text{Cov}_{t_i}(2,1) = \gamma^2\int_{t_{i-1}}^{t_i}(Q(t_i;\Theta) - Q(\tau;\Theta))d\tau + \omega_{t_{i-1}}^2\Delta Q_i, \quad (17)$$

$$\text{Cov}_{t_i}(2,2) = \gamma^2\int_{t_{i-1}}^{t_i}(Q(t_i;\Theta) - Q(\tau;\Theta))^2 d\tau + \eta_B^2\Delta t_i + \omega_{t_{i-1}}^2\Delta Q_i^2.$$

Then the posterior distribution of $\varphi_{t_i}$ given $Z_{t_i}$ (also the conditional distribution of the multi-dimensional Gaussian distribution) can be inferred as $\varphi_{t_i} \sim N(\psi_{t_i}, \omega_{t_i})$. Its expectation and variance are respectively given by

$$\psi_{t_i} = \psi_{t_{i-1}} + \text{Cov}_{t_i}(1,2)\text{Cov}_{t_i}(2,2)^{-1}(Z_{t_i} - \psi_{t_{i-1}}Q(\tau;\Theta)), \quad (18)$$

and

$$\omega_{t_i} = \text{Cov}_{t_i}(1,1) - \text{Cov}_{t_i}(1,2)\text{Cov}_{t_i}(2,2)^{-1}\text{Cov}_{t_i}(2,1). \quad (19)$$

Each time a new observation becomes available, the distribution of $\varphi$ will be updated by Eqs. (18)-(19).

### III. DATASET PROCESSING

To demonstrate the effectiveness of the proposed approach, we conduct numerical experiments on NASA C-MAPSS (Commercial Modular Aero-Propulsion System Simulation) turbofan engine degradation data [24]. The C-MAPSS data set is divided into four subsets (FD001, FD002, FD003, FD004) according to the operational conditions and engine serial numbers. Each of these subsets consists of multivariate time series collected from 21 sensors, but the operation settings are only available for FD002 and FD004. The four subsets have 100, 260, 100, and 248 samples, respectively, with RUL distributions illustrated in Fig. 7. One can see that FD001 and FD002 have similar RUL distributions, while the distributions of FD003 and FD004 have heavier tails. To test the robustness of our model under different operational conditions, we first estimate the model parameters only with FD002, and then apply the trained model to all the datasets.

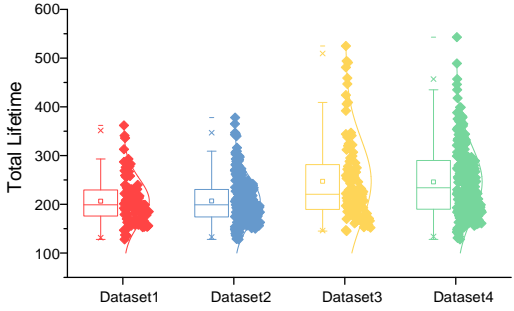

Fig. 7. RUL distributions of the four subsets in C-MAPSS.

### A. Health Index Construction

The Health Index (HI) is constructed via stacking [25]. Stacking is an effective operation in achieving better predictive performance and mitigating overfitting by combining information from multiple predictive models to generate a new model. The stacking operation involves two steps. Different regressors are used to extract the base HIs in the first step, and the second step treats these HIs as features for the final HIs. In this study, sensor data s2, s3, s4, s11, s17, and the running cycles are chosen as the input features, and their corresponding labels in the training set are constructed by normalizing the running cycles engine by engine, i.e.,

$$N(t_i^j) = \frac{t_i^j - \min_j(t_i^j)}{\max_j(t_i^j) - \min_j(t_i^j)}, \quad (20)$$

where $t_i^j$ is the running cycles of engine $j$ at time $i$. The range for the HIs is [0, 1]. The XGB Regressor, Extra Trees Regressor, and the Random Forest Regressor are chosen as base models, while the final HI is constructed by averaging their predictions. 80% of the sensor data in FD002 are used as the training set, while the remaining 20% are used for testing. Their corresponding HIs (constructed by stacking models) are shown in Fig. 8. The failure threshold is chosen as 0.95. The specialty of the dataset lies in that the HIs show different behavior before and after a diffusion point, where a dispersion phenomenon is observed. That explains why in some studies RULs are considered as a constant initially, to omit the nuance in the sensor data [26]. However, it would improve the prediction accuracy at the early age dramatically if the model is able to recognize the hidden pattern in the initial period.

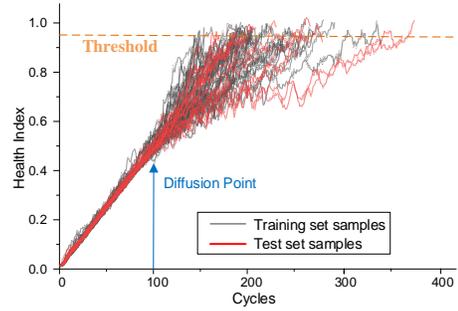

Fig. 8. HI for the training and test samples in FD002.

### B. Evaluation Metrics

Several performance metrics are employed to evaluate the performance of the proposed approach, focusing on two critical indicators: (a) point estimation of mean values, and (b) degree of uncertainties. The first criterion is usually measured by RMSE, and the second can be characterized by the coverage probability of the prediction interval (PICP) or the mean prediction interval width (MPIW). PICP measures if the actual RUL is within the predictive confidence interval (CI), which is given by

$$\text{PCIP} = \frac{1}{n} \sum_{i=1}^{n} rul_i \quad (21)$$

where $rul_i = 1$ if $L_i \leq RUL_i \leq H_i$ and 0 otherwise; $H_i$ and $L_i$ are the upper and lower bounds of the CIs. MPIW is the mean width of the CI. The PICP and the MPIW can be conflicting metrics, as a narrower MPIW will certainly reduce PICP.

## IV. EXPERIMENTAL RESULTS

In this section, we compare the performance of the proposed adaptive DNN with methods from the literature such as non-adaptive DNN, adaptive Wiener process, and non-adaptive Wiener process. The Wiener-based methods use the basic linear model as the coexistence of acceleration and deceleration, as shown in Fig. 8. The adaptive DNN and the non-adaptive DNN have the same network structure. More specifically, in the LSTM-CNN structure the encoder has one recurrent layer with 20 hidden nodes and a dropout probability of 0.3. The decoder CNN has two intermediate 1-dimensional convolutional layers, with 16 and 32 filters each. The kernel sizes for the two layers are both 50. The final layer is used to map the intermediate outputs into a one-dimensional vector of 400 step predictions. Dropouts of 0.3 are also applied on each layer of the CNN decoder. The Adam optimizer with a learning rate of 0.0015 and decay of 2e-5 is used for model training.

### A. Parameter Estimation

The parameters of the adaptive Wiener and the adaptive DNN model are summarized in Table II, including the initial drift rate $\varphi_{t_0}$, diffusion coefficient $\gamma^2$, and noise degree $\eta_B^2$.

TABLE II
PARAMETER ESTIMATION

| Method | $\varphi_{t_0}$ | $\gamma^2$ | $\eta_B^2$ |
|---|---|---|---|
| Adaptive Wiener | 4.67e10-3 | 1.16e-3 | 7.00e-6 |
| Adaptive DNN | 1 | 6.36e-4 | 6.41e-6 |

The estimated $\varphi_{t_0}$ for the linear model is 4.67e10-3, while it is assigned as 1 for the adaptive DNN. The coefficients $\gamma^2$ and

$\eta_B^2$ can be viewed as the measurements of epistemic and aleatoric uncertainties, respectively. Table II states that the two models share similar degrees of aleatoric uncertainties (i.e., similar values for $\eta_B^2$). Such uncertainty emerges from random factors and is hard to track. In contrast, the epistemic uncertainty of the adaptive DNN model, also known as the model uncertainty, is much smaller than that of adaptive Wiener model due to the ability of DNN to extract knowledge from historical information. Accordingly, the total uncertainties are decreased.

### B. Model Performance on FD002

Experiment is first conducted on test set FD002. Note that update of the drift parameter $\varphi$ takes place every 10 cycles of sensor data and no update in the first 20 cycles of data. Table III summarizes the average values of RMSE, PCIP (90% CI), and MPIW, and RMSEs are graphically shown in Fig. 9. The proposed method (adaptive DNN) outperforms adaptive Wiener, and adaptive DNN improves over non-adaptive DNN only insignificantly from 17.71 to 16.25 in RMSE value. This is due to the fact that the training and test data sets of FD002 show similar degradation pattern (hence, not much to learn by being adaptive).

TABLE III
MODEL PERFORMANCE ON FD002

| Method | RMSE | PCIP | MPIW |
| --- | --- | --- | --- |
| Adaptive DNN | 16.25 | 99.45 | 81.27 |
| DNN | 17.71 | - | - |
| Adaptive Wiener | 43.22 | 99.45 | 122.22 |
| Wiener | 45.62 | 100.00 | 118.41 |

Fig. 9 shows the RMSEs of the four prognostic approaches. All four methods show relatively flat RMSE values in the first 100 cycles (not much learning) before reducing RMSE values afterwards. Throughout the experiment, the DNN-based models outperform Wiener-based models significantly.

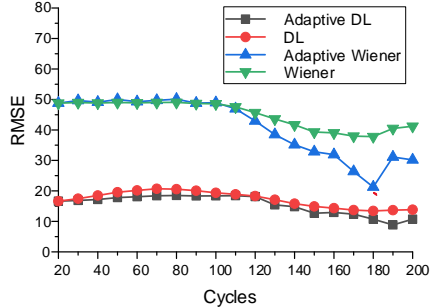

Fig. 9. The RMSEs of the four methods on FD002

### C. Model Performance on the other datasets

We further apply the model trained on FD002 to datasets FD001, FD003 and FD004, which were collected under different operational conditions. The prediction accuracies are reported in Tables IV-VI, and RMSE values are graphically shown in Fig. 10. Notice in Table IV-VI that MPIW is influenced by both the predicted trajectory and the confidence interval (CI). Even MPIWs of Wiener processes could be small, the corresponding RMSEs are still too large.

Adaptive DNN shows significantly better performance than adaptive Wiener, and the improvement made by the adaptive DNN (Wiener) over non-adaptive DNN (Wiener) is more noticeable than the earlier experiment done with only FD002. In general, prediction errors increase from the previous experiment as expected: all the methods were trained with FD002 and applied to FD001, FD003 and DF004 without re-training. Importantly, the adaptive version of DNN and Wiener show noticeably better performance than non-adaptive versions, testifying the value of online learning given unseen data.

TABLE IV
MODEL PERFORMANCE ON FD001

| Method | RMSE | PCIP | MPIW |
| --- | --- | --- | --- |
| Adaptive DNN | 22.44 | 99.88 | 68.94 |
| DNN | 28.25 | - | - |
| Adaptive Wiener | 47.77 | 100.00 | 93.21 |
| Wiener | 49.10 | 100.00 | 93.69 |

TABLE V
MODEL PERFORMANCE ON FD003

| Method | RMSE | PCIP | MPIW |
| --- | --- | --- | --- |
| Adaptive DNN | 32.74 | 100.00 | 183.79 |
| DNN | 47.97 | - | - |
| Adaptive Wiener | 82.94 | 100.00 | 193.33 |
| Wiener | 94.82 | 100.00 | 122.19 |

TABLE VI
MODEL PERFORMANCE ON FD004

| Method | RMSE | PCIP | MPIW |
| --- | --- | --- | --- |
| Adaptive DNN | 21.12 | 98.68 | 148.79 |
| DNN | 32.59 | - | - |
| Adaptive Wiener | 69.72 | 98.79 | 182.43 |
| Wiener | 82.84 | 98.79 | 113.68 |

A careful review of Fig. 10 offers useful observations. First, all four methods show similar performance patterns between FD002 (Fig. 9) and FD001 (Fig. 10-(a)) and between FD003 (Fig. 10-(b)) and FD004 (Fig. 10-(c)), which makes a good sense given similarity of RUL distributions between FD001 and FD002 and between FD003 and FD004 as shown in Fig. 7. Second, the DNN-based methods outperform the Wiener-based methods with all four data sets. Third, the adaptive versions of both DNN-based and Wiener-based methods outperform their non-adaptive cousins. More interestingly, the gap is larger with FD003 and FD004, which have significantly different RUL distributions from FD001 and FD002. Notice that all four methods are initially trained with FD002. That is, the learning effect is more significant when the test data is more different than the training data. Fourth, with all four data sets, the effect of learning starts insignificant and grows.

Figs. 11-12 illustrate the prognosis done by the adaptive DNN and the non-adaptive DNN methods on two particular cases: case 103 from FD002 and case 10 from FD003. Note that $\varphi Q(t)$ is the prognostic done by the adaptive DNN model, while $Q(t)$ by the non-adaptive DNN model. In Fig. 11, prediction trajectories, the CIs, and the corresponding RUL distributions at cycles 50, 100, and 150 are shown, whereas in Fig. 12 the same indicators are shown later times at cycles 100, 200, and 300 considering longer lifetime of the engine. Both the adaptive and the non-adaptive DNN models in Fig. 11 shows accurate prediction. Interesting to note that the proposed adaptive DNN model keeps its estimates of $\varphi$ close to 1,

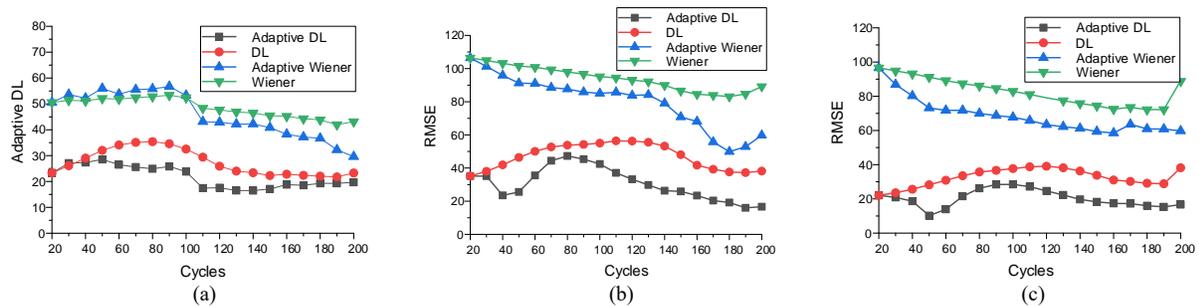

Fig. 10. The RMSEs of the four methods on the three datasets: (a) FD001; (b) FD003; and (c) FD004.

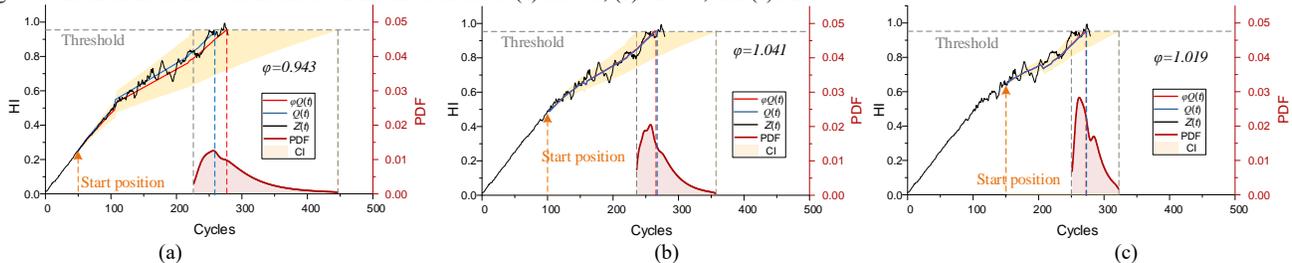

Fig.11. The prognosis results for case 103 in FD002. (a) from 50 cycles; (b)from 100 cycles; (c) from 150 cycles.

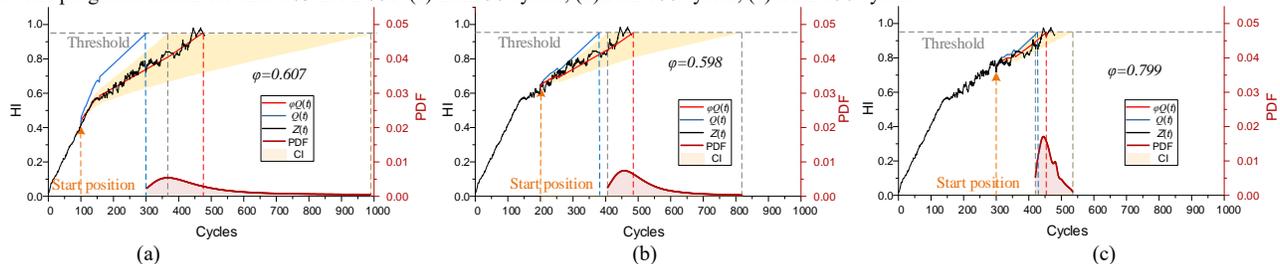

Fig.12. The prognosis results for case 10 in FD003: (a) from 100 cycles; (b) from 200 cycles; and (c) from 300 cycles.

implying little need to adaptively control the drift rate. In Fig. 12, however, the adaptive DNN model estimates the drift rate away from 1, hinting the non-adaptive DNN model shows poor accuracy. In both Fig. 11 and Fig. 12, our model successfully predicts the degradation pattern well before inflection point, at which degradation pattern dramatically changes as shown in Fig. 8. On the other hand, the non-adaptive DNN model overshoots the progression of degradation at the 100[th] and the 200[th] cycle in Fig 12-(a) and Fig 12-(b), and to lesser degree at the 300[th] cycle in Fig. 12-(c). This demonstrates the necessity of being adaptive via learning the drift rate. By adjusting the drift parameter, the hybrid model can adapt to new working conditions, while maintaining its historical knowledge.

## V. CONCLUSION

We proposed a novel prognostic framework integrating Wiener-based degradation processes with deep neural network. Instead of presetting a degradation trajectory, we learn the evolution pattern from historical health data via a powerful DNN structure, namely, LSTM-CNN. To the best of our knowledge, this is the first Wiener-based degradation model which is capable of both learning and self-adaptive. This significantly enhances the robustness and flexibility of the prognostic framework. Furthermore, our proposal can capture both aleatoric and epistemic uncertainties, and effectively alleviate the generalization problem of DNN approaches via online updating. The superior performance of our proposed adaptive DNN to the standard non-adaptive DNN model and the traditional Wiener-based approaches has been validated in numerical experiments conducted on NASA turbofan engine degradation data.

The current work has two possible extensions. First, other stochastic processes, such as random coefficient and Gamma, can be employed as the degradation model. Second, the single-variate prognostic framework can be extended to multi-variate frameworks.